# Predicting County Level Corn Yields
# Using Deep Long Short Term Memory Models

**Zehui Jiang**

Iowa State University

gingerzh@iastate.edu

**Chao Liu**

Tsinghua University

cliu5@tsinghua.edu.cn

**Nathan P. Hendricks**

Kansas State University

nph@ksu.edu

**Baskar Ganapathysubramanian**

Iowa State University

baskarg@iastate.edu

**Dermot J. Hayes**

Iowa State University

dhayes@iastate.edu

**Soumik Sarkar**

Iowa State University

soumiks@iastate.edu

**Abstract**

Corn yield prediction is beneficial as it provides valuable information about production and prices prior the harvest. Publicly available high-quality corn yield prediction can help address emergent information asymmetry problems and in doing so improve price efficiency in futures markets. This paper is the first to employ Long Short-Term Memory (LSTM), a special form of Recurrent Neural Network (RNN) method to predict corn yields. A cross sectional time series of county-level corn yield and hourly weather data made the sample space large enough to use deep learning technics. LSTM is efficient in time series prediction with complex inner relations, which makes it suitable for this task. The empirical results from county level data in Iowa show promising predictive power relative to existing survey based methods.

## 1. Introduction

In the 2001 Nobel Prize winning paper "The Market for Lemons"[1], George Akerlof shows that in a second-hand car market where asymmetric information exists, sellers know the quality of their cars but the buyers do not. Buyers offer a price based on the expected quality of the car. Sellers of the high-quality cars which are worth more than the average price will exit the market. This drives up the proportion for low value cars and in turn drives down the offer price. Eventually, only "lemons" are left in the market and the market collapses. The key to this collapse is that sellers have more information than buyers. This is called information asymmetry.

The solution to information asymmetry is to provide public information to all participants in the market at the same time. The United States Department of Agriculture (USDA) has been predicting national corn yield and production every year since 1964. The main method they use is survey based. They also use enumerators who make field visits in important corn production areas. The results from this traditional statistical method is subjective since it is the farmer's estimation at the point the survey was taken. USDA has tried new sources of data collection such as satellite imagery from MODIS (moderate resolution imaging spectroradiometer) from NASA. However, as of fall 2017, USDA continues to rely on the survey-based data.

Several private companies are possibly in a position to improve on the USDA survey. Examples include Lanworth[2], Tellus Labs[3] and Climate Corp[4]. These companies set up plant growth models based on weather information and expert knowledge and they monitor satellite imagery and weather patterns. They incorporate as many independent lines of evidence as possible into their estimates and produce daily yield estimates in contrast with the monthly state-level prediction from the USDA. However, customers need to pay these companies for their estimates, possibly resulting in information asymmetry in the market. Traders in the CME corn futures who have preferential access to this information may be in a position to make profitable trades to the detriment of traders who do not have access. The motivation for this work is to utilize modern data science to provide public daily corn yield prediction at a county-level and to aggregate this information to a national level. The long run intent is to eliminate information asymmetry in corn futures markets. Our work indicates that modern machine learning method has the potential to improve predictions relative to the USDA.

*1.1 Background Knowledge*

Corn is mainly planted in the Midwestern part of the United State (the green area in Figure 1), in an area called the Corn Belt. The region is characterized by level land, deep fertile soils, and a high organic soil concentration[5]. USDA usually reports the nationwide corn yield in late February of the year following harvest. Corn yield growth increased rapidly after 1950 with genetic improvement in seed and in farm management (Figure 2).

Past research using machine learning predicted yield with discrete weather variables[6,7]. To the best of our knowledge, no research has been done to predict yield at any point during the growing season using only data accumulated up to that point. To accomplish this we use Long Short-Term Memory (LSTM)[8], a special form of Recurrent Neural Network (RNN)[9,10,11].Its efficiency in capturing long-term dependencies and predicting time series with complex inner relations makes it a perfect choice for our work. Though LSTM is one of the most popular methods in deep learning, it has never been used in any other field except natural language processing. This work is the first to apply LSTM in crop yield prediction, and demonstrates its potential to address other prediction problems.

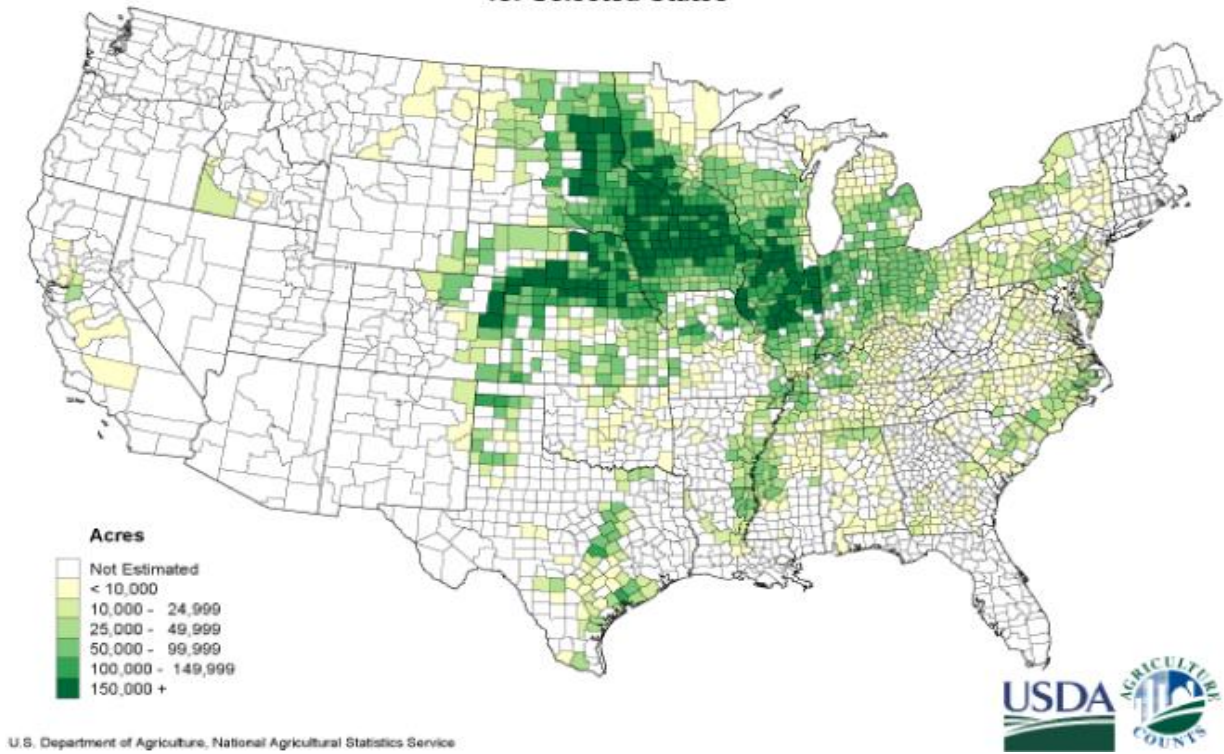

Figure 1: Corn Belt in the United States

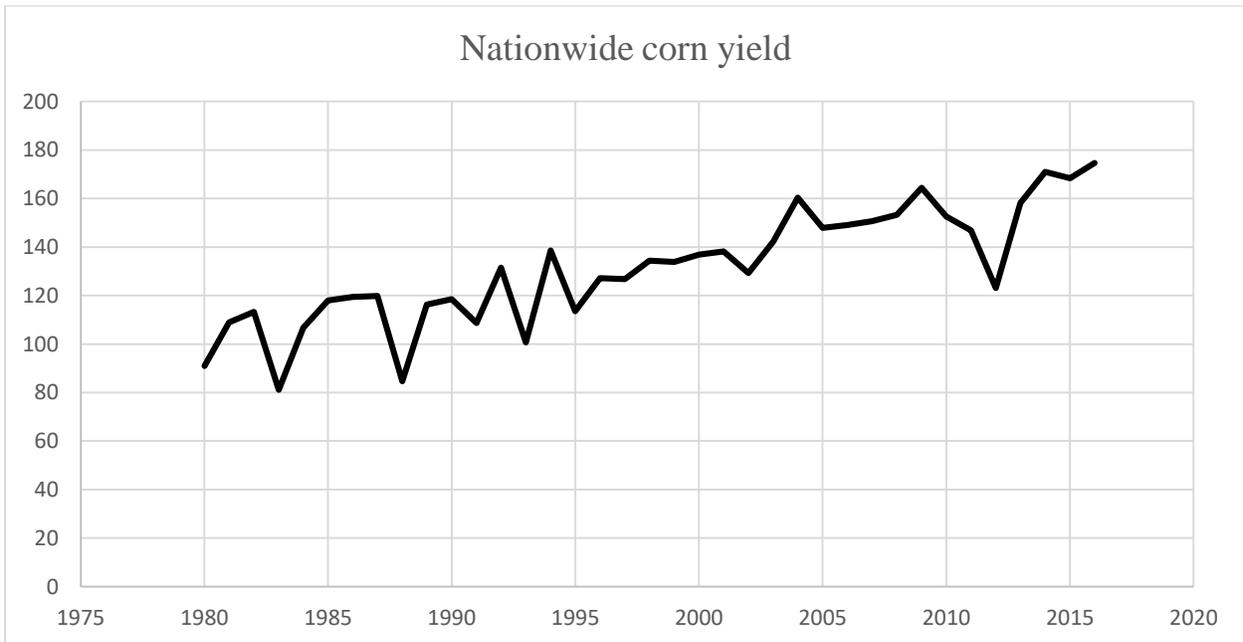

Figure 2: Nationwide corn yield from 1980 to 2016

## 2. Literature Review

When building reasonable prediction models, expert knowledge is an important input variable in selection and preprocessing. Dr. Fred Below completed experiments in corn growth and published "The Seven Wonders of the Corn Yield World" in 2008[12]. Weather wins the first place among the seven "wonders". Rain, temperature, wind and humidity are shown to be key yield determinants.

Foot and Bean (1951) were the first to provide evidence of trends and patterns in crop yields associated with weather[13]. Kaylen and Koroma (1991) suggest limiting weather variables to temperature and precipitation to model U.S. corn yields[14]. They present a linear model using a stochastic trend and monthly rainfall and temperature variables during May to August. Deschenes and Greenstone (2007) find that yield decreases in temperature and increases in rainfall[15]. All the above literature assumes linear relationship between corn yield and weather variables.

Schlenker and Roberts (2009) conclude that temperature has a nonlinear effect on corn yield[16]. They used nationwide county level data and show a steep non-linear decline in yields when temperature is above 29°C. Tian Yu et al. (2011) examine the drought effect on crop yield in Iowa, Illinois, and Indiana and find significant results[17]. They also estimate non-linear weather impacts on corn yield using the Bayesian approach.

Machine learning has been considered to predict crop yields[18,19]. Monisha and Robert (2004) use an artificial neural network (ANN) model with rainfall data to predict corn and soybean yield[6]. The Maryland Water Quality Improvement Act of 1998 requires mandatory nutrient management planning on all agricultural land in Maryland. In order to predict yields under typical climatic condition, they choose a machine learning method and find ANN models consistently produced more accurate yield predictions than regression models. Newlands and Townley-Smith (2010) were the first to apply Bayesian Network (BN) into crop yield prediction[20]. They attempt to predict energy crop yield and provide predicted probability distributions. However, they do not attempt to provide point estimates since BN is designed for categorical variables.

## 3. Data Collection

### 3.1 Yield Data

County level data was collected for each of Iowa's 99 counties from 1980 to 2016. The first 33 years were selected as training data, while the most recent four years were used for out of sample forecasting. Historical observed Iowa county yields were downloaded from National Agricultural Statistics Service (NASS) Quick Stats[21]. There are a total 37*99=3,663 records of yield data and 3,267 training samples. Corn yields increase through time due to genetic gain. Therefore, we need to adjust the historical corn yield into the same base. Li (2014) indicated that genetic gain for corn in Iowa was almost 2.5 bu/ac per year from 1980 to 2000 and 4.67 bu/ac per year after 2000[22]. We used Li's results and applied arithmetic and geometric (1.5% increase per year) yield changes to de-trend the yield data.

### 3.2 Weather & Soil Data

Three types of input variables associated closely with corn yield are available. They are hourly weather data, soil quality data and soil moisture data. The hourly weather data was purchased from a professional weather data company – Weather Underground[23]. The data is a representative of an entire $19 \times 19$ mile area, which is more accurate than the commonly used weather station data which measures weather at one central point. We use weather data from month of April to October to reflect the growing season in the Corn Belt.

Rainfall in the growing season may result in high yields, but flooding that results in standing water can reduce yield. High wind speed will damage corn crops by uprooting plants and can increase evapotranspiration. The maximum, minimum and the mean temperature all influence yields. Industry experts use a concept called Growing Degree Days (GDDs)[24] to predict plant development rates. GDD are calculated by

$$GDD = \frac{T_{\max} + T_{\min}}{2} - T_{base}$$

where $T_{\max}$ = min(86ºF, daily maximum temperature).

$T_{\min}$ = max(50ºF, daily minimum temperature).

And $T_{base}$ is the base temperature required to trigger the optimum growth. It equals 50ºF for corn. Accumulated GDD during the growing season is an important factor.

Soil moisture has a critical impact on corn yield. The Palmer Drought Severity Index (PDSI)[25] is a long term cumulative measure of water availability in the soil. It is a standardized index that spans -10 (dry) to +10 (wet). 0 stands for a normal moisture condition, negative shows the soil is dry and positive means there is surplus water. PDSI uses temperature data and a physical water balance model to capture the basic effect of global warming on drought. PDSI is monthly data downloaded from the National Oceanic and Atmospheric Administration (NOAA) at the Crop Reporting District (CRD) level (Supplementary Figure 3). We match the counties with each district and assign the value to those counties (i.e. counties in the same CRD has the same PDSI value).

Soil quality[26] data was collected from the SSURGO database[27] (database for storing gridded soil survey results) and aggregated to the county level using only areas classified as cropland according to the NLCD[28] (National Land Cover Database). For continuous variables, we aggregated to the county level by taking the average. The data covers the whole Corn Belt region at county level. There are over one hundred variables in this dataset. Each variable is a constant number for each county, since soil quality typically does not change over time. We pick fourteen variables from the data, which we think are most related to corn yield (Figure 1). Root zone for water storage and soil droughty vulnerability are considered the most two significant soil variables.

*3.3 Variable selection and data preprocessing*

The time series of input variables can be expressed in hourly or daily format. Each county for each year is a sample record with output value — yield, and the corresponding input time series falling into the growth period April to October, so the length of the input time series $\{x_t\}$ would be t=5,136 for hourly input vectors or t=214 for daily input vectors. However, for hourly inputs vectors, there are too many parameters needed to estimate with only 3,267 training samples. Therefore we use a daily input sequence $\{x_t\}$ with t=214.

**Variable Selection**. There are totally twenty-eight candidate input variables. In addition to the fourteen soil quality variables and PDSI, we also include max/min/mean temperature for each day, total daily rainfall, daily average wind speed, max rainfall during the day, accumulated rainfall and GDD accumulated up to that date. Since July is the most important month for corn growth[29], rainfall and max temperature in July are also included. The ratio of acres planted for corn divided by the total acres planted may also influence the corn yield since farmers in corn intensive counties will specialize in corn management techniques and management. Two interaction terms (max temperature*soil droughty and max*PDSI) are also included. The idea behind this is that at high temperatures soil moisture will be more important than at low temperatures.

First, we trained the model with all 28 variables included. Our county level yield predictions turned out to be almost a horizontal line at the county average yield level. We believe this was due to the use of too many constant terms which is irrelevant to the model. We use MRMR[30] (minimum redundancy maximum relevance), a feature selection method introduced in Peng (2005) to rank the soil quality data and eliminated those variables that had low rank, we also calculated the correlation between weather variables and eliminated the highly correlated ones. After trial and error with different combination of the input variables and with expert knowledge, we arrived at the "best" ten input variables for corn yield prediction. These are max/min/mean temperature, total daily rainfall, wind speed, soil root space for holding water (rootznaws), soil droughty vulnerability (droughty), PDSI, accumulative rainfall and GDD.

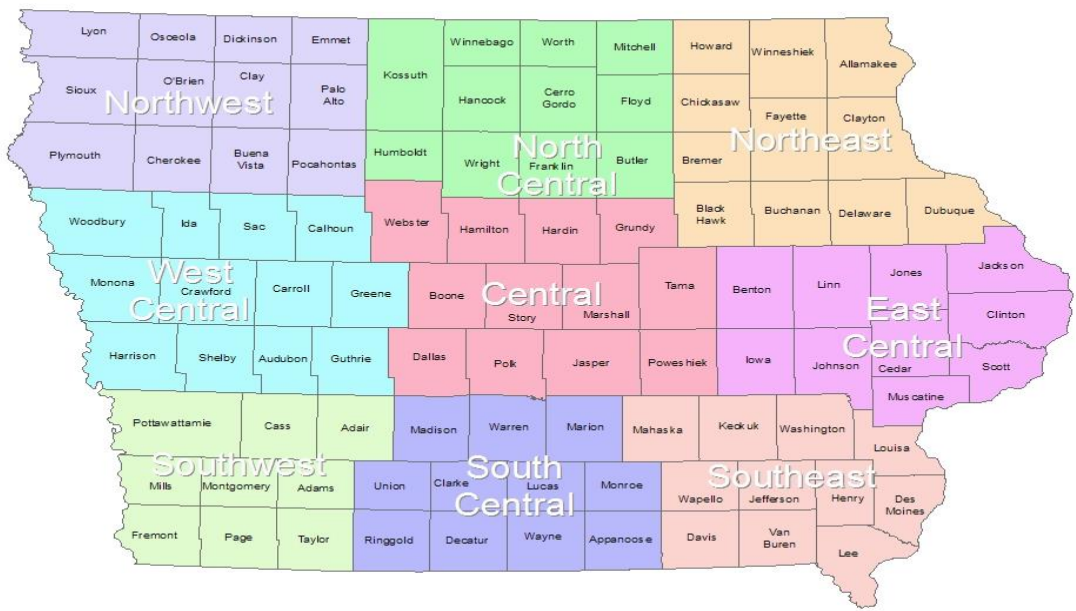

U.S. Department of Agriculture Crop Reporting Districts

Figure 3: Iowa Crop Reporting District Map

**Data Augmentation**. Even when using daily as opposed to hourly input series, 3,267 training samples are still not enough. In order to generate more training samples we pick two or three counties from the same CRD (Crop Reporting District, see Figure 3) in Iowa, and take the average of their yield and input variables respectively, then a new sample is created. There are nine CRD in Iowa, hence the total number of training samples added with combination samples increases up to 70,026. These combination samples should be reasonable since PDSI is also collected at the CRD level and all other data are also average numbers for county area (the most precise data point should be each farmland, which is not available).

Ten input variable sequences are stored in the format of 3D tensor cube. This is a key step to make our data fit into the model. All input data should be converted into tensor format for LSTM training in computer. Figure 1c is an example of 3D tensor cube. X-axis indicates the number of the input variables, Y-axis is the length of the time series and Z-axis shows the number of samples. Hence the dimension of our 3D tensor is $10 \times 214 \times 70026$. PDSI is monthly data, so it repeats the times of number of days in each month. While rootznaws and droughty repeat 214 times since they are constant.

## 4. Methodology

A supervised method – recurrent neural network (RNN)[9,10,11] – is used first. RNN is a family of neural networks for processing sequential data. It is typically used in text prediction and speech recognition. RNN is very useful for cases where there are nonlinear and unknown interactions but it does not provide causal relationships. The challenge is to fit our problem into RNN format. Even though RNN is specially created for time series data, previous applications have focused the prediction of the following points in the same time series. This paper develops a novel way adapt RNN to predict crop yields a problem where crop yield responses from prior decades may have predictive power.

### 4.1 Recurrent Neural Network

For a regular RNN, the network consists of three layers: an input layer, a hidden layer and an output layer. $x^{(t)}$ is the input sequence, $y^{(t)}$ is the output sequence and $h^{(t)}$ is a series of hidden states. The number of the hidden layers is not constrained to one. In the deep learning recurrent neural networks, the number of the hidden layers can reach eight or more. Adding hidden layers can help to study the more complex structure of the model, but also requires more data. $U, V, W$ are shared weights that we need to learn. And there is an activation function $f$ that $h^{(t)} = f(Ux^{(t)} + Wh^{(t-1)})$, which should be chosen before learning the networks. The corn yield prediction problem could not fit into a regular RNN, therefore we use the many-to-one RNN model here (Figure 5). Many-to-one RNN model is suitable when there is sequence input with one output, thus it is perfectly match with our data format described in the data section.

*4.2 Long short-term memory (LSTM)*

The mathematical challenge of learning long-term dependencies in recurrent networks is called the vanishing gradient problem. As the length of input sequence increases, it becomes harder to capture the influence of the earliest stages. The gradients to the first several input points vanish and become equal to zero. Therefore a special RNN model called Long short-term memory (LSTM)[8] is developed. In the recurrency of the LSTM the activation function is the identity function with a derivative of 1.0. So, the backpropagated gradient neither vanishes or explodes but remains constant. Figure 5 shows the difference in the framework between regular RNN and LSTM, where tanh (hyperbolic tangent function) is a commonly used activation function. It is clear that LSTM has a more complex structure to capture the recursive relation between the input and hidden layer.

LSTM adds a new sequence $\{c_t\}$, called cell state, upon the regular RNN. Cell state is a space that is specifically designed for storing past information, i.e. the memory space. It mimics the way the human brain operates when making decisions. Operation is executed to update old cell state $c_{t-1}$ to $c_t$. This is the time where we actually drop old and add new information. We can get the output as $h_t$, which is the same process as regular RNN.

a

| Variables | Explanation |
|---|---|
| ffd | number of days between the last freezing temperature and the first freezing |
| sandtotal | total sand, sand means mineral particles 0.05mm to 2.0mm in diameter |
| silttotal | total silt, silt means mineral particles 0.002mm to 0.05mm in diameter |
| claytotal | total clay, clay means mineral particles less than 0.002mm in diameter |
| om | weight percentage of organic matter (decomposed plant and animal residue) |
| bulkDensity | the oven dry weight of the less than 2 mm soil material |
| lep | linear expression of the volume difference of soil fabric and oven dryness |
| caco3 | the quantity of Carbonate (CO3) in the soil expressed as CaCO3 |
| ec | the electrical conductivity of an extract from saturated soil paste |
| soc0_150 | soil organic carbon stock estimate (SOC) in standard layer |
| rootznaws | root zone available water storage (mm) |
| droughty | soil droughty vulnerability determined by earthy major components |
| sand | percentage of sand contained in the soil |
| share_cropland | cropland share of the whole county land |

b

| Number | Input variables | Selection criterion |
|---|---|---|
| 10 | Mean/max/min temperature, rainfall, wind speed, PDSI, rootznaws, droughty, accumulative GDD and rainfall | mRMR, expert knowledge and trial and error |
| 15 | Mean temperature, rainfall, wind speed, PDSI, rootznaws, droughty, accumulative GDD, acre_share, frost free days, total clay, organic matter, electrical conductivity, max rainfall, rainfall in July, max temperature in July | Correlation matrix |
| 16 | All input variables excluding the twelve soil quality variables except rootznaws and droughty | mRMR |
| 28 | All available input variables included | all input variables |

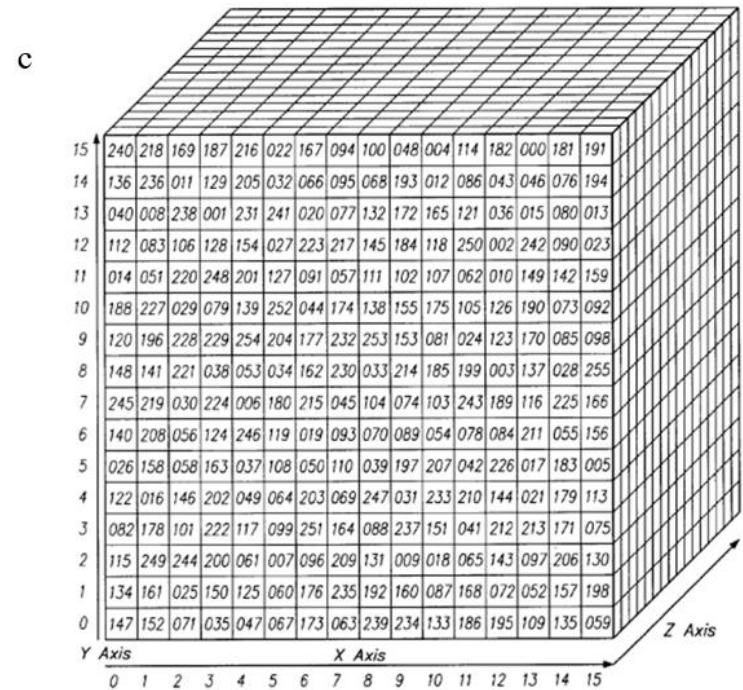

a shows the detail for soil related candidate variables.
b shows the process for variables selection
c shows how the data are stored for training

Figure 4: Details about variables description, variable selection and data preprocessing

Figure 5: Model Representation

*4.3 Training of LSTM*

The algorithm to learn recurrent neural network is Stochastic Gradient Descent (SGD)[31] and back-propagation through time (BPTT)[32]. Backpropagation is the most widely used algorithm in the training of multi-layer neural networks. The core idea behind BP is the composite function chain rule. The loss function we pick here is the mean squared error. SGD is an efficient algorithm to searching for the local minimum of the loss function. Then BPTT algorithm is used to compute the gradient for the equation $h^{(t)} = f(Ux^{(t)} + Wh^{(t-1)})$ and the loss function. The nodes of our computational graph include the parameters $U, V, W$ and constant terms as well as the sequence of nodes indexed by $t$ for $x^{(t)}$ and $h^{(t)}$. Once the gradients on the internal nodes of the computational graph are obtained, we can obtain the gradients on the parameter nodes. The parameters are shared across time steps. Given a starting point, calculating the gradient of that point, and searching in the direction of negative gradient. This is the fastest way we search for a local minimum. Then we could update the parameters with iteration of the SGD by searching for smaller local minimum.

Our LSTM model has been learned using a Python package called 'keras'[33] on top of Theano[34] backend. Hyperparameter searching is an important process before the commencement of the learning process. We assign a set of numbers for hyperparameters such as learning rate, number of hidden layers, number of hidden nodes in each layer, dropout rate and let the machine randomly pick one value in the set for each hyperparameter. Usually after searching for over 300 models with different combination of hyperparameter settings, we can find the 'best' model and the corresponding 'best' hyperparameters.

## 5. Results and Discussion

Our choses model has two hidden layers, daily input vectors and two combination samples. Figure 6 and 7 shows the "best" prediction with percentage and constant adjustment respectively.

The black line is our prediction while the red is the true yield. There is also a chart showing the absolute error between the prediction result and the true yield. We include the absolute error as a standard to judge the performance of the model.

Our initial model starts with hourly input vector $\{x_t\}$ where t=5,136 and yield adjusted to both 2013 and 2015 base with 1.5% yearly increase. Number of samples are 3,267 and there is one hidden layer. The input variables are hourly temperature, rainfall, wind speed, PDSI, soil root space for holding water, soil droughty, accumulative rainfall and GDD by hour. Figure 8 shows the prediction results for this original model. The left column shows the results for yield adjusted to a 2013 base while the right is to 2015. After comparing the results between the left and right part, we conclude that whichever year the yield is adjusted to, there will be hardly any influence on prediction results. Therefore, we uniformly adjust the yield to 2013 base for all models.

We train the initial model with daily input vectors and try to improve the model from three changes in the settings: 1. use two hidden layers instead of one; 2. adding more training samples created with combination method (two+three counties average); 3. including more input variables (variable selection from the 28 variables introduced in section 3.3). Table 1 shows that 10 "best" input variables with 2 hidden layers and combination samples has the smallest mean squared error (Figure 9). Nevertheless, the fluctuation of prediction is still less than the true yield. Does this average trend exist because of including too much combination samples? Therefore, we trained the "best" model again with only two counties combination samples added, which totals 19734 samples. We also use the constant genetic gain adjustment with yield data.

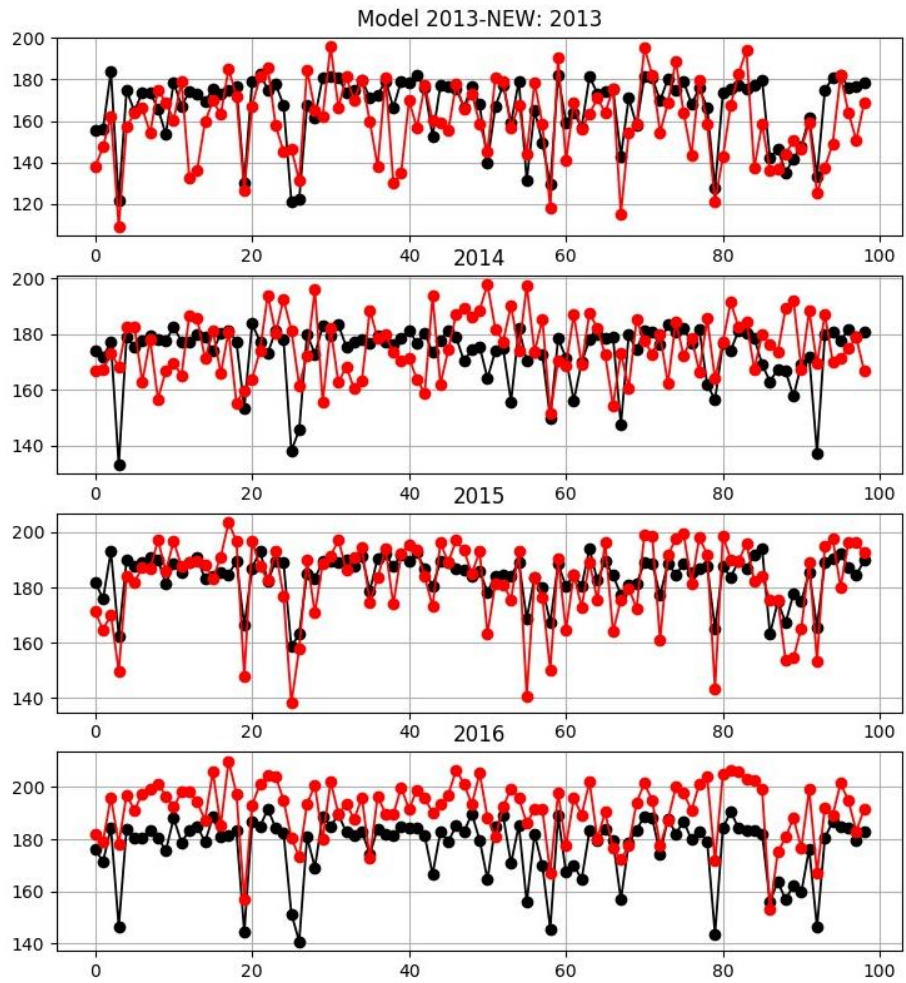
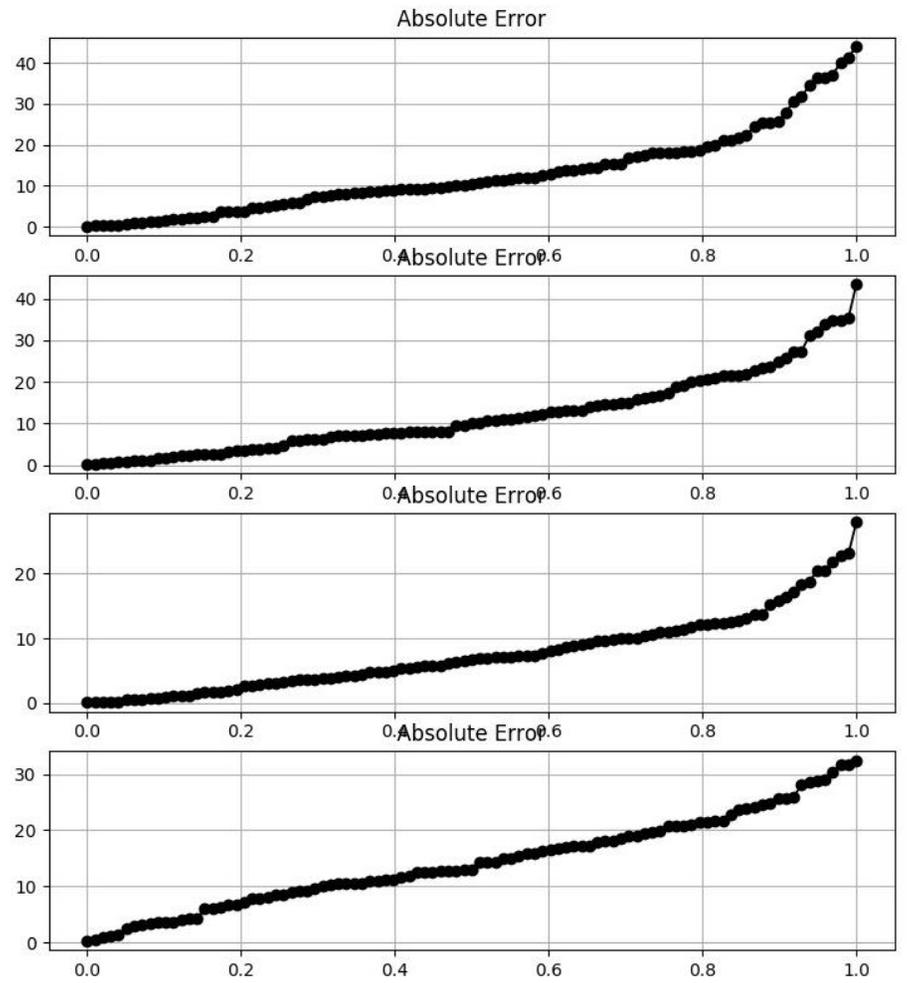

Figure 6: Prediction Results for the "Best" LSTM model with percentage adjustment

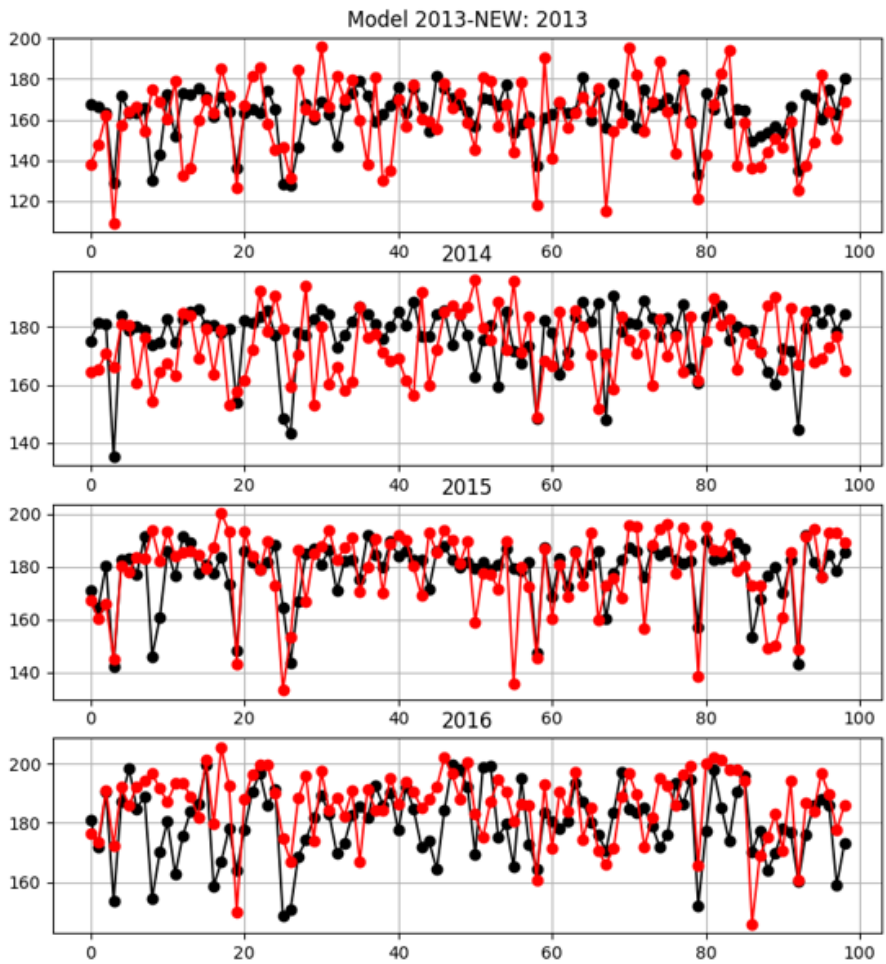
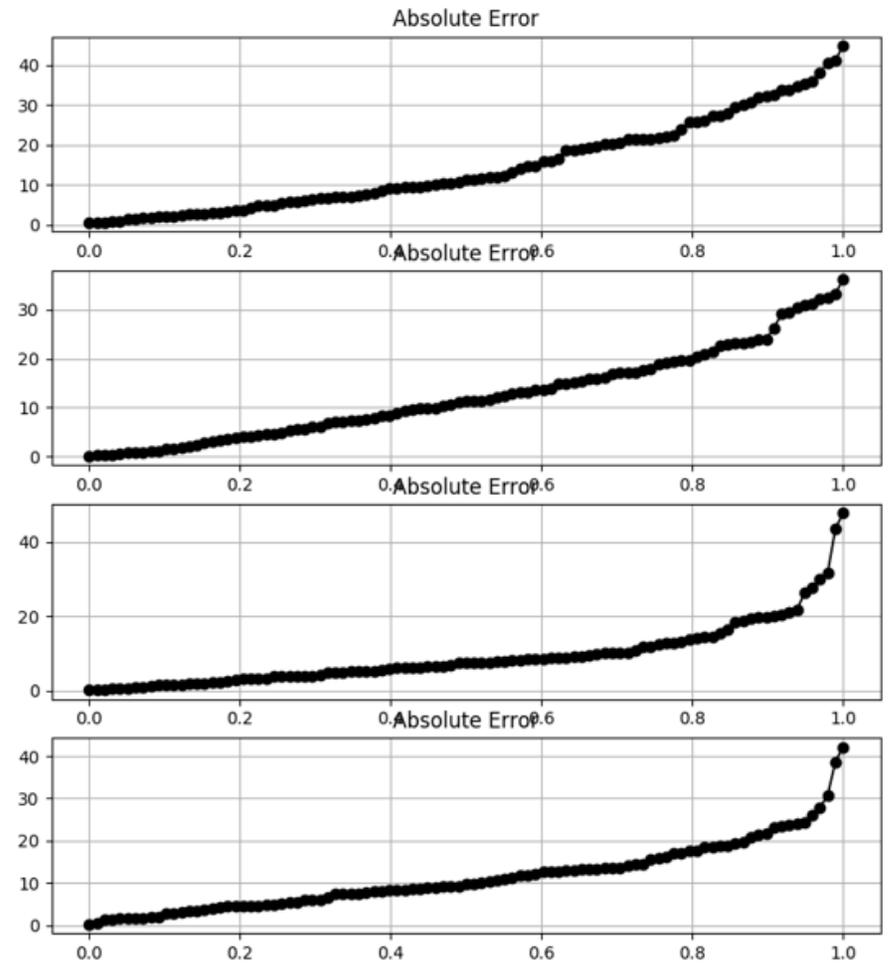

Figure 7: Prediction Results for the "Best" LSTM model with constant adjustment

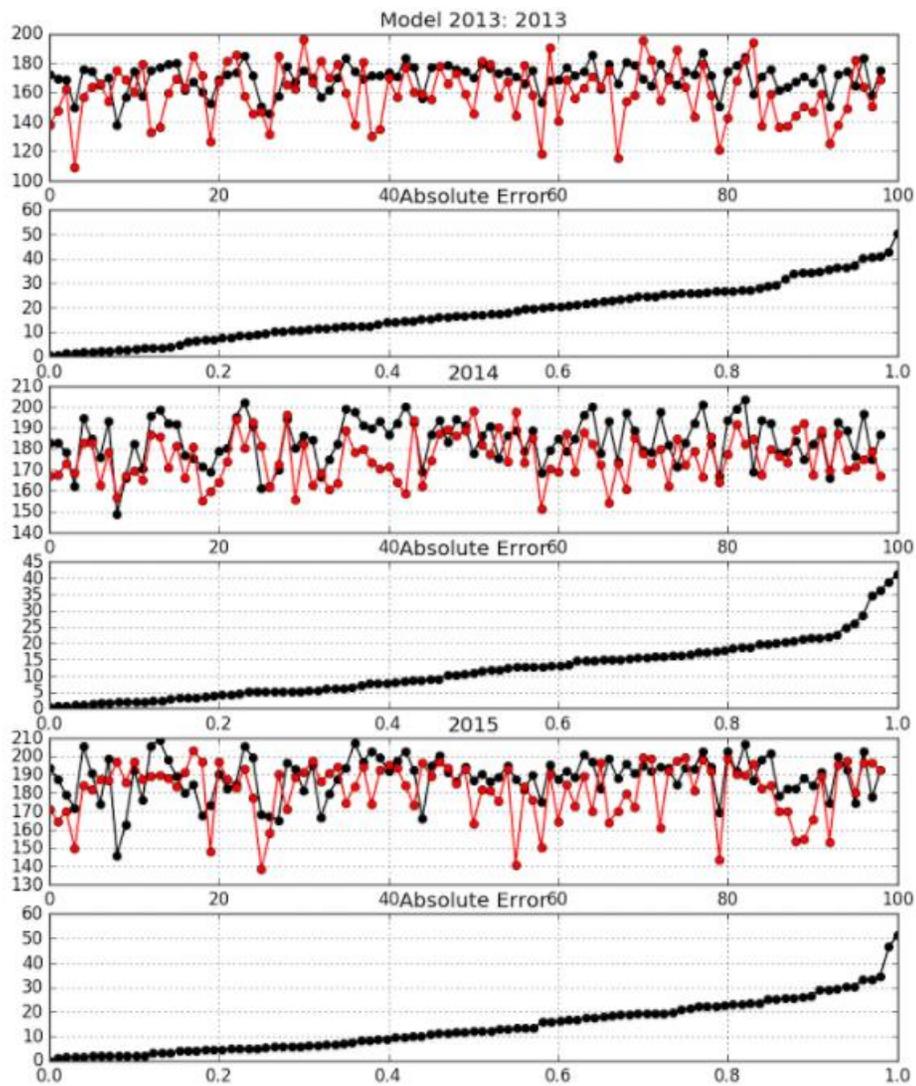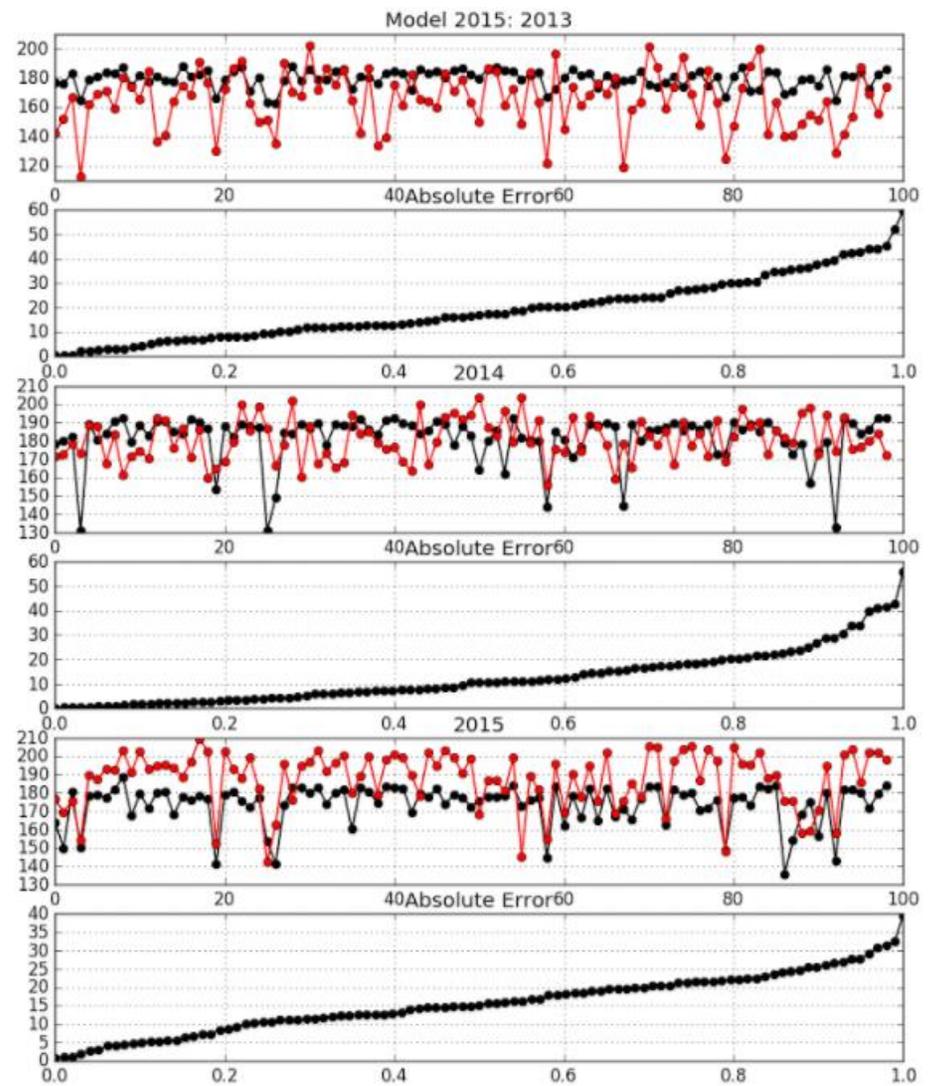

Figure 8: Prediction Results for hourly input vectors with initial model

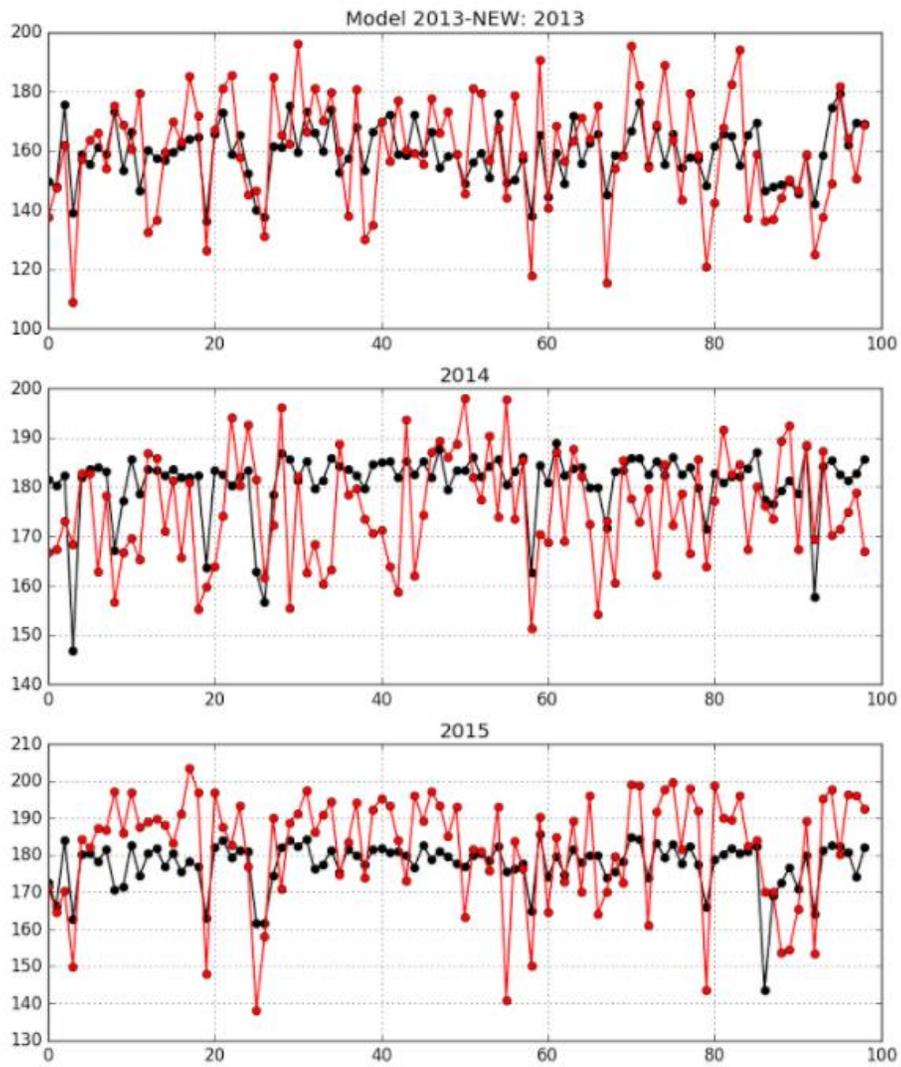
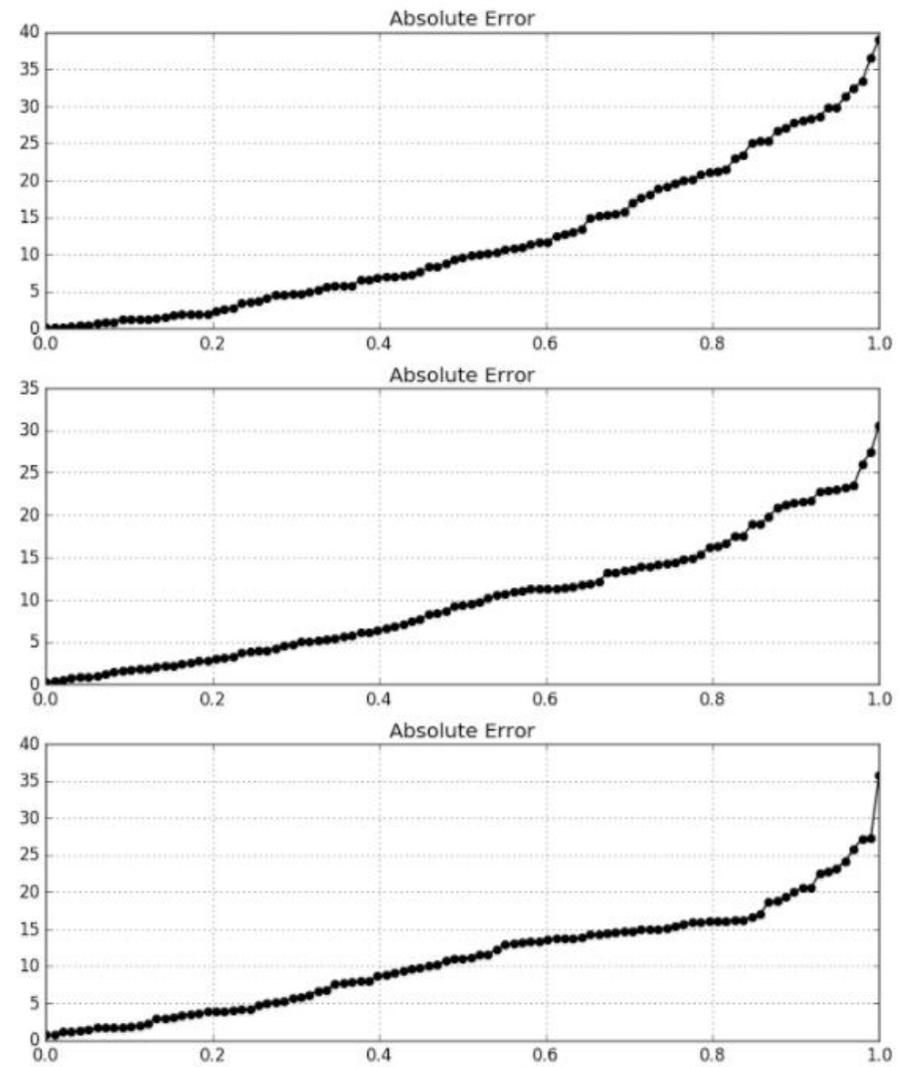

Figure 9: Prediction Results for two hidden layers LSTM with combination samples

| Number of input variables | Number of samples | Hidden layers | Mean squared error |
|---|---|---|---|
| 10 | 3267 | 1 | 255.404 |
| | 70026 | 1 | 211.6265 |
| | 3267 | 2 | 233.2547 |
| | 70026 | 2 | 191.0535 |
| 15 | 70026 | 2 | 361.9132 |
| 16 | 70026 | 2 | Very large |
| 28 | 70026 | 2 | Very large |

**Table 1: Comparison of different LSTM model settings**

| year | yield | USDA | 1 layer | 1 layer com | 2 layers | 2 layer com |
|---|---|---|---|---|---|---|
| 2013 | 164.00 | 169 | 157.59 | 160.80 | 165.39 | 161.15 |
| 2014 | 178.00 | 183 | 185.27 | 180.21 | 177.62 | 185.12 |
| 2015 | 192.44 | 189 | 188.26 | 185.39 | 180.65 | 184.71 |

**Table 2: Comparison of state level predictions of different LSTM model settings**

| year | yield | USDA | percentage adjustment | constant adjustment |
|---|---|---|---|---|
| 2013 | 164.00 | 169 | 171.33 | 165.57 |
| 2014 | 178.00 | 183 | 179.13 | 184.13 |
| 2015 | 192.44 | 189 | 192.22 | 190.50 |
| 2016 | 203.04 | 199 | 189.13 | 195.45 |

**Table 3: Comparison of state level predictions with two "best" models**

Table 2 and 3 compares the prediction results at the state level. USDA regularly reports their state level yield prediction every August, September, October and November. Here we compare with their November and final prediction against ours. For year 2013, constant adjustment has the best prediction, for year 2014 and 2015, percentage adjustment perfectly predict the yield while USDA and constant adjustment has almost the same performance. However, for 2016, our prediction are much lower than the actual yield, which indicating that the weather event in 2016 were not represented in the historical data.

### 7.1 Early Prediction

In order to make comprehensive comparison between USDA and our model, we also trained other three models as early prediction for data available to August, September and October respectively (i.e. y=122, 153 and 183 for the 3D tensor cube in section 3.3). Figure 10 summaries the prediction results of USDA and our LSTM model. All of our models are trained with almost 700 hyperparameters' sets, which means an optimal model should have been reached. This comparison indicates the power of our model and the ability to beat USDA with limited data.

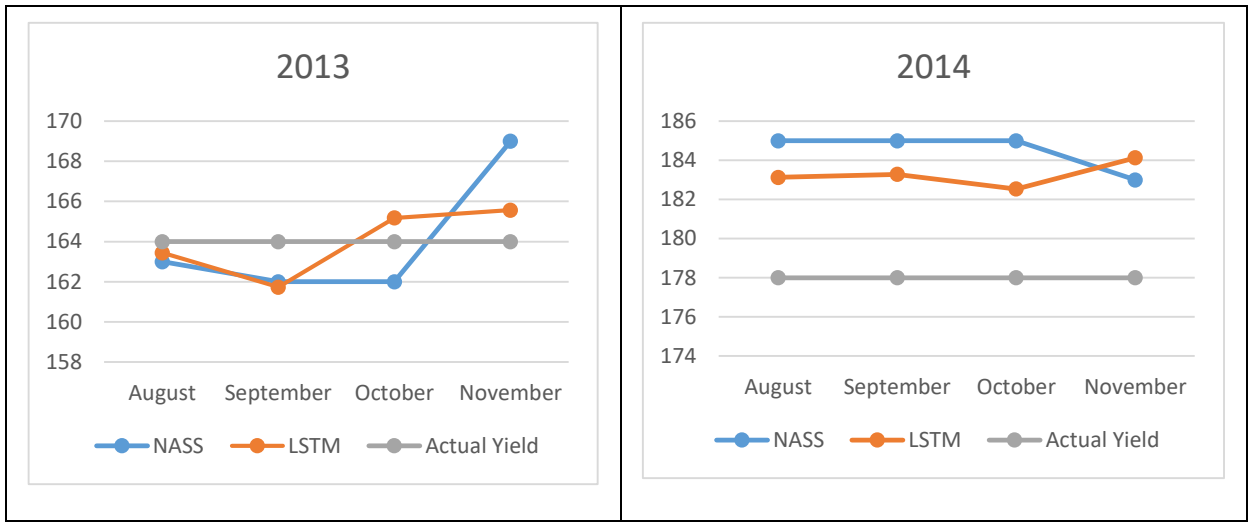

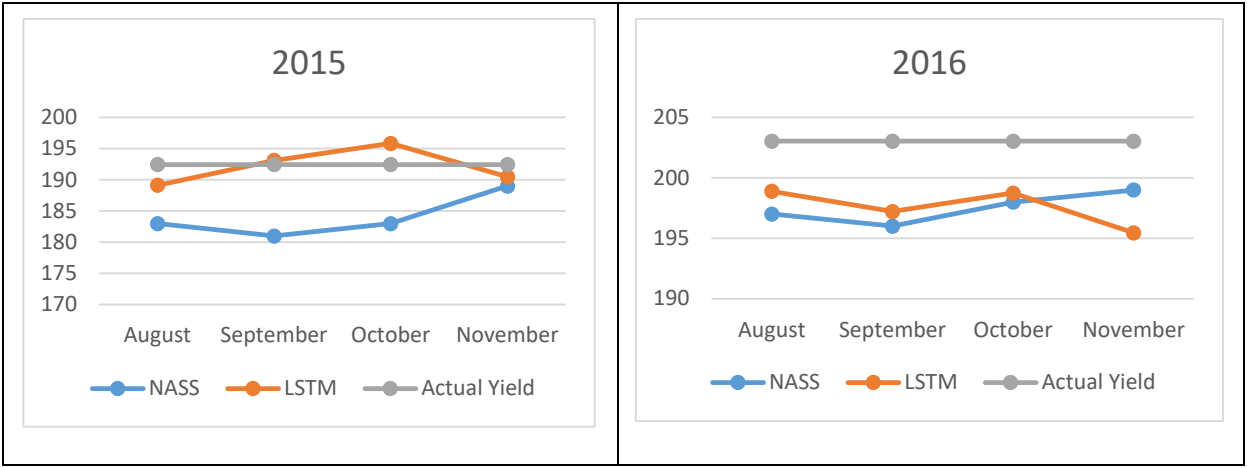

Figure 10: Yield Prediction Comparison between USDA and LSTM

## 6. Summary and Future Work

In this paper, we demonstrate a LSTM model for county-level corn yield prediction in Iowa. The model performs reasonably well based on expert's opinion, which indicates the potential of LSTM in yield prediction rather than in language processing. The good performance of early monthly prediction through LSTM has shown the possibility of high-quality daily prediction which is publicly available. While the main future goal of this research is to provide accurate daily county-level prediction for the whole Corn Belt, we also consider to design specific models for prediction under extreme cases such as flood or drought in the future.